\documentclass[lettersize,journal]{IEEEtran}
\usepackage{amsmath,amsfonts}
\usepackage{algorithmic}
\usepackage{algorithm}
\usepackage{array}
\usepackage[caption=false,font=normalsize,labelfont=sf,textfont=sf]{subfig}
\usepackage{textcomp}
\usepackage{stfloats}
\usepackage{url}
\usepackage{verbatim}
\usepackage{graphicx}
\usepackage{cite}
\usepackage{booktabs}
\usepackage{adjustbox}
\usepackage{multirow}
\usepackage{xcolor}

\begin{document}

\title{Alleviating Regional Shortcuts for Few-Shot Class-Incremental Learning}

\author{Haichen Zhou*, Yazhe Lyu*, Yixiong Zou,~\IEEEmembership{Member,~IEEE}, Ruixuan Li,~\IEEEmembership{Member,~IEEE}, Yuhua Li,~\IEEEmembership{Member,~IEEE}
\thanks{*These authors contributed equally to this work.}
\thanks{The authors are with the School of Computer Science and Technology, Huazhong University of Science and Technology, 
Wuhan 430074, China. Emails: 
m202273832@hust.edu.cn,
yazhelv@hust.edu.cn,
yixiongz@hust.edu.cn,
rxli@hust.edu.cn,
idcliyuhua@hust.edu.cn. }
\thanks{Corresponding author: Yixiong Zou.}
}

\markboth{Journal of \LaTeX\ Class Files,~Vol.~18, No.~9, September~2020}%
{How to Use the IEEEtran \LaTeX \ Templates}


\maketitle

\begin{abstract}
Few-shot class-incremental learning (FSCIL) aims to incrementally learn novel classes with only a few samples while avoiding forgetting base classes.
However, current methods show a tendency to misclassify novel-class samples into base classes, which we find to be caused by the excessive focus on base-class-discriminative regions on novel-class samples. In this work, we aim to explore the underlying mechanism for an interpretation and solution. We first provide a compositional view to analyze the transferred and reused spatial patterns on novel-class samples. Then, through extensive experiments and theoretical analysis, we identify both empirically and theoretically that a shortcut exists in the model's base-class training, which naturally forms the excessive focus on only the most discriminative regions (primitives), which we term as the \textbf{regional shortcut}.
Finally, based on this interpretation, to address this problem, we propose a compositional-learning-based method to learn two primitive sets (a common set and a discriminative set), which alleviates the regional shortcut by constraining the model to learn and utilize the common primitive set for base- and novel-class recognition.
Extensive experiments on standard FSCIL benchmarks demonstrate the effectiveness of our approach, yielding consistent improvements over existing state-of-the-art methods in both accuracy and interpretability.
\end{abstract}


\section{Introduction}
\IEEEPARstart{F}{ew}-Shot Class-Incremental Learning (FSCIL) is proposed to handle the continuously arising real-world knowledge with only limited data\cite{wang2022learning,tao2020few}, aiming to first learn from abundant data in the base session (base classes), and then transfer the knowledge to downstream novel sessions (novel classes) with scarce training data.
The main challenges in FSCIL arise from two aspects: catastrophic forgetting\cite{hinton2015distilling} of learned knowledge (base classes), caused by the sequential nature of the incremental learning task, and overfitting to novel classes, due to the scarcity of training data. 

\begin{figure*}[!t]
\centering
\includegraphics[width=\textwidth]{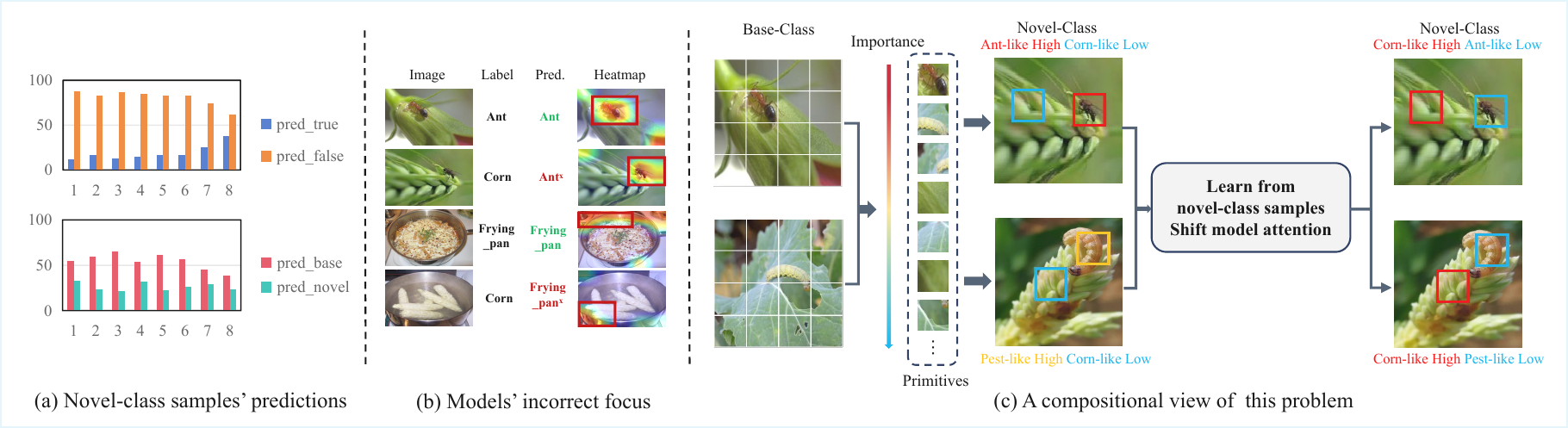} 
\vspace{-0.5cm}
\caption{
    (a) FSCIL models tend to make wrong predictions on novel-class samples (pred\_false), where the wrong predictions as base classes (pred\_base) are much more than those as novel classes (pred\_novel). 
    (b) We find the cause is in models' incorrect focus on base-class-related discriminative regions in novel-class data, e.g., only focusing on the ant-like insect on the corn and misclassifying a corn (novel class) as an ant (base class), albeit the given novel-class training samples.
    (c) We are inspired to provide a compositional view to delve into this problem, which decomposes base-class knowledge into primitives (patches) and transfers them to novel classes for knowledge re-composition.
}
\vspace{-0.5cm}
\label{intro}
\end{figure*}

To address these challenges, a common strategy is to freeze the feature extractor pretrained on base classes during novel-class learning\cite{ zhang2021few,wang2023few}, which stably shows good performance despite its simple design. However, 
it also leads to limited adaptation and performance on novel classes. 
Specifically,
we first find the model tends to classify a substantial portion of novel-class samples into base classes (Fig.\ref{intro}(a)), revealing a bias in the model’s decision. While this has been observed across FSCIL benchmarks, the mechanistic reasons behind it remain insufficiently studied.

To understand this behavior, we visualize the heatmap of those misclassified samples, and we observe that the model tends to incorrectly focus on a small number of discriminative regions associated with base classes. For example, focusing on the ant-like insect on the corn and misclassifying a corn (novel class) as an ant (base class), as in Fig.\ref{intro}(b).
Although it is natural to focus on the learned base-class pattern on novel classes, the model should gradually shift its attention to the corn by learning from the novel-class training samples, albeit of their scarcity. However, we find that the model shows very low activation on corn-like patterns across most training samples, which makes it extremely difficult to learn to recognize such patterns.
In other words, we hold that such a bias in learned and transferred base-class spatial patterns limits the adaptation to novel classes.

To address this problem, in this paper, we are inspired by recent compositional learning works\cite{zou2024compositional} for an interpretation, which analyzes from the aspect of decomposition and re-composition of base-class spatial patterns. These works typically view image patches as visual primitives, which decompose base-class knowledge and are transferred into novel classes for knowledge re-composition, as in Fig.\ref{intro}(c). In other words, ideally, the model should transfer both the ant-like primitives and corn-like primitives to novel classes, and shift its attention to corn-like primitives by seeing more corn-like samples.
However, quantitative and qualitative studies such as Fig.\ref{intro}(b) show that the model only captures ant-like patterns (base-class-discriminative primitives, BCD primitives) but totally ignores other visual patterns, leading to the incorrect focus on the ineffective regions of novel-class samples.
Moreover, we find both empirically and theoretically that such an excessive focus on BCD primitives is inherited from the model's inherent shortcut in learning the most discriminative regions, which we term as \textbf{regional shortcuts}.

Based on this interpretation, we propose a novel compositional method to mitigate biased primitives caused by regional shortcuts. Specifically, we introduce an additional common primitive set that captures class-agnostic, shared semantic cues across categories. This encourages the model to learn more transferable and semantically enriched primitives during the base-class training phase, thereby alleviating misclassification on novel classes that arises from over-reliance on BCD primitives and neglect of all other ones. Notably, by visualizing learned primitives, our method shows better interpretability of the models’ behavior.
Our contributions can be listed as:
\IEEEpubidadjcol
\begin{itemize}
\item We find that current FSCIL models tend to misclassify novel classes by incorrectly focusing on discriminative regions associated with base classes, reflecting a bias in the learned and transferred spatial patterns, which limits the model’s adaptation to novel classes.
\item We provide a compositional view into this problem, and interpret it both empirically and theoretically as an excessive focus on BCD primitives caused by the model's regional shortcuts during base class training.
\item Based on this interpretation, we further propose a compositional method that incorporates a common primitive set, encouraging the model to learn more transferable and semantically enriched primitive representations, thereby reducing over-reliance on BCD primitives and neglect of all other primitives.
\item Extensive experiments validate the rationale of our interpretation and method, and show that we can consistently outperform state-of-the-art works.
\end{itemize}

\setlength{\textfloatsep}{8pt}

\section{Related Work}
\label{sec:related}
\textbf{Few-Shot Class-Incremental Learning} (FSCIL) \cite{hou2019learning,zou2022margin} aims to learn novel classes with few examples per incremental session. Current FSCIL challenges mainly stem from catastrophic forgetting \cite{french1999catastrophic} and overfitting to novel classes \cite{zhu2021self}.
Our research targets FSCIL challenges, especially novel-class performance. TEEN \cite{wang2023few} offers a training-free prototype calibration based on semantic similarity to reduce misclassifying novel as base classes. RDI \cite{zhou2024delve} links poor novel-class performance to label-irrelevant redundancies, suggesting decoupling and adding a dummy class to improve the feature space.
Recent FSCIL studies further address these challenges from different perspectives, including generative co-memory regularization~\cite{bao2026few}, prior-knowledge infusion~\cite{bao2025pki}, constrained dataset distillation~\cite{bao2025cd2}, static-dynamic collaboration~\cite{bao2025divide}, attraction redistribution~\cite{zhao2025attraction}, and language-guided relation transfer~\cite{zhao2024language}. These methods mainly focus on memory preservation, prior knowledge transfer, distilled sample construction, or cross-modal relation modeling to improve incremental adaptation.
Unlike prior approaches, we conduct an in-depth investigation into why models confuse novel and base classes, attributing the performance degradation on novel classes to the excessive focus on base-class discriminative regions during the processing of novel-class samples. We are the first to attribute this confusion to regional shortcuts from base-class training, differing from RDI's view. Based on it, we propose a novel method to boost performance and interpretability.

\textbf{Compositional learning}, inspired by cognitive theories~\cite{biederman1987recognition}, aims to generalize to novel combinations by leveraging the compositional structure inherent in many visual and linguistic concepts. This principle has been widely applied in zero-shot learning\cite{li2024context} and few-shot learning\cite{zou2024compositional}, where methods often decompose visual categories into attribute–object pairs\cite{ bao2024prompting} or part-based representations\cite{weng2023decompose}, enabling the recognition of unseen compositions by recombining known primitives.
Recent works aim to learn semantically expressive and spatially comprehensive primitive sets to better capture the structure of visual concepts. Disentangled representation methods\cite{wang2024disentangled} enhance interpretability and generalization but often lack explicit constraints on spatial diversity. Attention-based compositional models\cite{zheng2024layer} aggregate multi-region information to reduce reliance on overly discriminative features, yet they typically treat all primitives equally without distinguishing shared from distinctive elements. Prototype-based frameworks \cite{qu2025learning} encourage compact and meaningful representations, but seldom differentiate primitives by their semantic roles or levels of generality.
Unlike prior works such as~\cite{zou2024compositional}, which implicitly learn shared primitives, our method explicitly constructs two primitive sets and mitigates regional shortcuts by enforcing the use of a common set for both base and novel classes.

\begin{figure*}[!t]
\centering
\includegraphics[width=0.7\textwidth]{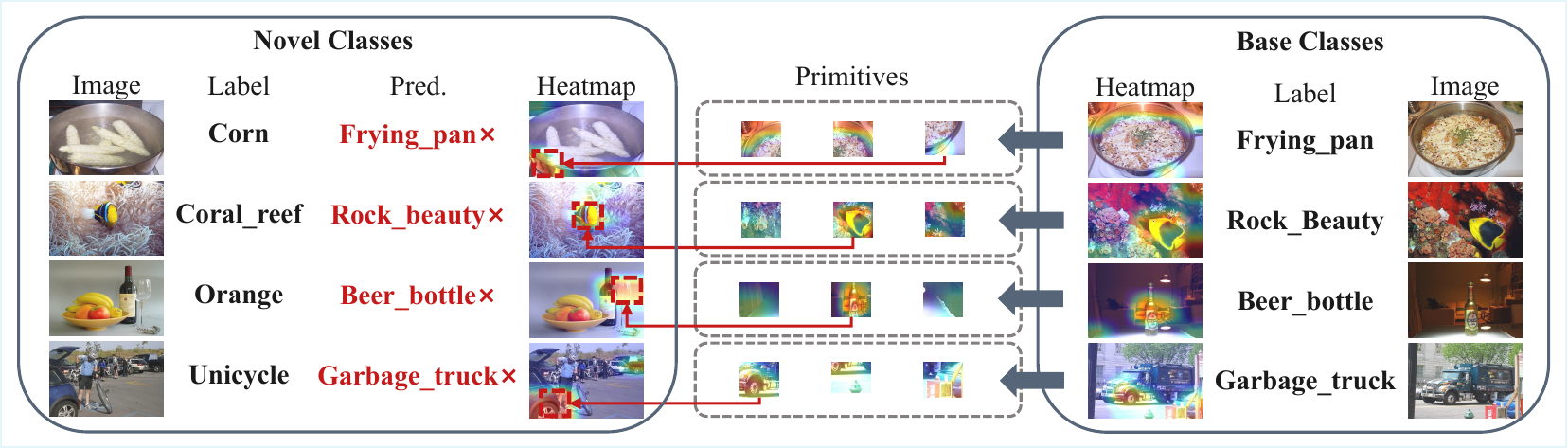} 
\vspace{-0.3cm}
\caption{
   More visualization of misclassified samples and transferred primitives for re-composition. 
}
\vspace{-0.5cm}
\label{wrong_pre} 
\end{figure*}

\section{A Closer Look at the Biased Spatial Patterns}
\label{sec:analysis}
In this section, we take a closer look at the biased spatial patterns in existing FSCIL models, which tend to focus on a small set of highly discriminative regions. While this behavior often boosts performance on base classes, it can significantly impair the ability to adapt to novel classes in few-shot incremental settings. From a compositional learning perspective, we analyze how visual primitives are learned, transferred, and recombined across sessions. Our analysis is based on experiments on miniImageNet, comprising 60 base and 40 novel classes across 8 incremental sessions. Through this perspective, we reveal the structural biases that limit effective generalization to novel classes and investigate the reasons behind misclassifying novel classes as base classes.

\vspace{-0.2cm}
\subsection{Preliminaries}
\label{protonet}
FSCIL continuously learns novel classes with a few examples. Specifically, the model learns classes in a sequence of $T$ training sessions denoted as $\left \{D_{train}^{0},D_{train}^{1},\ldots,D_{train}^{T-1}\right\}$, each associated with a distinct label space $\left \{C^{0},C^{1},\ldots,C^{T-1}\right \}$, ensuring no overlap between label spaces. Within each session, only the session-specific training set $D_{train}^{t}$ is accessible, comprising examples $\left \{ \left ( x_{i} ,y_{i} \right ) \right \}_{i=1}^{n_{t}}$ with $y_{i} \in C^{t}$ and $n_{t}$ denoting the number of examples. The testing set $D_{test}^{t}$, however, encompasses samples from all previously encountered and current classes, denoted as $\ D_{test}^{t}=\left \{ \left ( x_{i},y_{i} \right ) \right \}_{i=1}^{m_{t}}$, where $y_{i} \in \bigcup_{i=0}^{t}C^{i}$ and $m_{t}$ denoting the number of examples. The base session $(t=0)$ involves a training set with ample examples per class. Subsequent incremental sessions $(0 < t < T)$ consist of only limited samples. 

A common strategy is to fix the backbone trained on base classes while adapting to novel classes. During the base session, the model $f(x) = W^{T}\phi(x)$ is trained with:
\begin{equation}\label{soft}
\setlength{\abovedisplayskip}{5pt}
\setlength{\belowdisplayskip}{5pt}
		\mathcal{L}_{ce}=\frac{1}{N}  \sum_{i=1}^{N} -log \frac{e^{\tau cos(\phi(x_{i}),w_{y_{i}} )} }{e^{\tau cos(\phi(x_{i}),w_{y_i} )}+ {\textstyle \sum_{m\neq y_i}^{}e^{\tau cos(\phi(x_{i}),w_{m} )}} }, 
\end{equation}where $\phi(\cdot)$ represents the feature extractor and $W$ denotes the classifier for base classes. Then, $\phi(x)$ will be fixed and transitioned to the incremental session, and $W$ will be substituted with the novel-class classifiers.
Throughout the incremental session, the classifier for base classes remains fixed, while for novel classes, the classifier can be refined through fine-tuning or by employing the average embedding. During evaluation in session $t$, given an input \( x \), the predicted label \( \hat{y} \) is obtained by comparing the similarity between the feature \( \phi(x) \) and classifier weights \(\{w_k\}\) for all classes encountered so far:
\begin{equation}
\setlength{\abovedisplayskip}{5pt}
\setlength{\belowdisplayskip}{5pt}
    \hat{y} = \arg\max_{1 \le k \le \sum_{i=0}^t N_i} \langle \phi(x), w_k\rangle,
\end{equation}
where \( \phi(x) \in \mathbb{R}^d \) is the feature extractor output, \( w_k \in \mathbb{R}^d \) is the classifier weight vector of the \(k\)-th class, and \( N_i \) is the number of classes introduced in session \( i \).

\vspace{-0.6cm}
\subsection{Compositional View of FSCIL}
Humans possess a core cognitive ability to decompose complex information into simpler, meaningful components, enabling efficient learning and generalization by reusing familiar elements in new contexts\cite{hoffman1984parts}. Recent studies\cite{zou2024compositional} have achieved promising results based on compositional learning, which extracts visual primitives from previously acquired knowledge and recombines them to understand new concepts, where image patches are treated as candidate visual primitives. 
Inspired by it, we hold that the biased base-class spatial patterns observed in heatmaps (Fig.\ref{intro}(b)) can also be understood through compositional learning (Fig.\ref{wrong_pre}).

Following \cite{zou2024compositional}, let $\left\{X_{i}\right\}_{i=1}^{n}$ denote the set of feature representations extracted from all patches of an image, where the patch granularity follows the output resolution of the backbone, and let $\{Z_k^y\}_{k=1}^N$ represents the collection of patches corresponding to a specific class $y$. The cosine similarity computation between the image-level representation and class $y$ can thus be formulated as:
\begin{equation}
\setlength{\abovedisplayskip}{5pt}
\setlength{\belowdisplayskip}{5pt}
\begin{aligned}
\operatorname{sim}(\bar{X},\bar{Z}^y) 
&= \left(\frac{1}{n}\sum_{i=1}^n\frac{X_i}{\|X_i\|}\right) \cdot
   \left(\frac{1}{N}\sum_{k=1}^N\frac{Z_k^y}{\|Z_k^y\|}\right) \\
&= \frac{1}{nN}\sum_{i=1}^n\sum_{k=1}^N
   \frac{X_i}{\|X_i\|} \cdot \frac{Z_k^y}{\|Z_k^y\|}.
\end{aligned}
\end{equation}

From the above equation, we can see that the similarity between an image and a class $y$ can be reformulated as the average similarity across the set of image patches and the constituent elements (primitives) of class $y$. This suggests that the learning process of the baseline model can be viewed as decomposing an image into patches for primitive-level understanding, and then recomposing these patches to form new class representations for classification. Specifically, during the base stage, the model learns a set of primitives. In the incremental stage, when facing the challenge of limited samples for novel classes, the model generates semantic representations of novel classes by matching and composing primitives transferred from base classes for novel-class recognition.

\textbf{Ideal properties:}
Novel-class generalization is closely related to the re-composition of base-class primitives. 
Although it is natural for the model to reuse primitives that are most effective in base classes to represent novel classes, by encountering novel-class images, the model should gradually shift its attention to novel-class-discriminative primitives, even though these primitives are not the most discriminative ones in base classes. 
For example, in Fig.\ref{intro}(c), consider a base class "ant" and a novel class "corn", although the model still has the tendency to focus on the ant-like insect on the corn, by showing more corn images, the model should learn to increase the importance of corn-like patterns and decrease that of ant-like ones. This can be achieved by, for instance, averaging the primitive embeddings to form the novel-class prototype, i.e., since corn-like primitives are much more than ant-like primitives in novel-class images, their importance will naturally be higher.

However, in this example, this requires the model to at least capture corn-like primitives on base classes and transfer them to novel classes, although they may not be the most discriminative ones on base classes. In other words, during the base-class training, the model should \textit{learn a diverse primitive set instead of \textbf{only} focusing on the most discriminative ones}.
Therefore, we will study the re-composition and decomposition of primitives on novel and base classes, respectively.
\begin{figure*}[!t]
\centering
\includegraphics[width=6in]{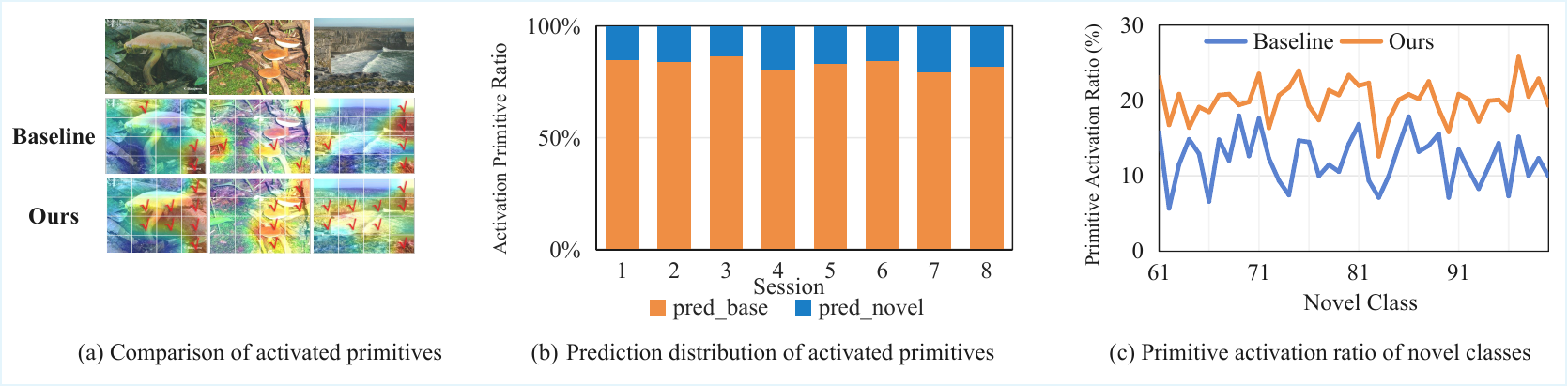} 
\vspace{-0.5cm}
\caption{
   (a) Activated primitives to show the re-composition on novel classes, where the baseline method can only activate (focus on) small regions that are incapable of representing novel classes. (b) Prediction of these attended regions for each novel-class image using the baseline model, which is heavily biased to base classes. (c) Quantitatively measuring the primitive activation ratio on all novel-class images. Combining with (b), it verifies that the baseline model classifies novel-class images primarily based on small regions closely associated with base classes, and can hardly focus on other useful regions.
}
\vspace{-0.4cm}
\label{novel_activation} 
\end{figure*}

\vspace{-0.4cm}
\subsection{Analysis from the Perspective of Composition Learning}
\subsubsection{\textbf{Novel-class re-composition}}
In Fig.\ref{wrong_pre}, we observe that the model's re-composition of primitives for novel-class samples is suboptimal, leading to misclassification. To gain deeper insights, we further analyze how the model activates its learned primitives for composition.
Following \cite{zhou2016learning}, we first formalize the criteria for a primitive to be considered \emph{activated}. Let the extracted feature map be $F \in \mathbb{R}^{d \times h \times w}$, where $d$ denotes the feature dimension and $h, w$ the spatial dimensions. Each primitive at location $(a,b)$ is represented by the vector $f(a,b) \in \mathbb{R}^d$. We compute an activation score $s(a,b)$ for each primitive and classify those exceeding a predetermined threshold as activated.

With this definition, we analyze predictions over the activated primitives (Fig.\ref{novel_activation}(b)).
The results show that most activated primitive compositions are misclassified as base classes, indicating a strong bias toward BCD primitives. This bias is exemplified in Fig.\ref{novel_activation}(a), where the model intuitively attends to regions closely related to base classes while overlooking other informative components
To verify this intuition, we further quantitatively verify it on all novel-class images (Fig.\ref{novel_activation}(c)), where we can see that the baseline model consistently exhibits a low activation ratio on novel classes, suggesting the model can hardly focus on other useful regions.
These observations suggest that the model over-relies on BCD primitives and neglects other informative components, resulting in representations biased toward base classes.

\subsubsection{\textbf{Base-class decomposition}}
\begin{figure*}[!t]
\centering
\includegraphics[width=\textwidth]{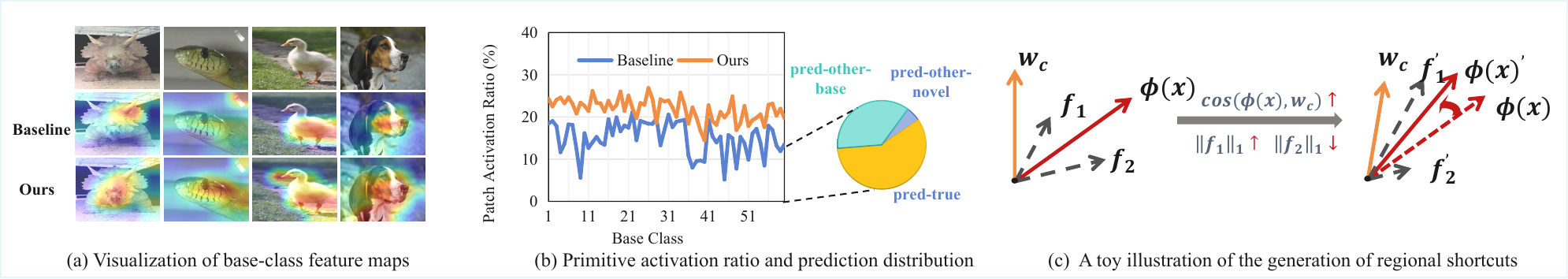} 
\vspace{-0.7cm}
\caption{
   (a) The baseline model tends to only focus on the most discriminative regions/primitives for base classes, which we term as \textit{regional shortcuts}. (b) Primitive activation ratio on \textit{base-class} images (left) and Prediction distribution of the most attended regions for each base-class image (right), verifying the attended primitives are the most discriminative ones for base classes. (c) To maximize $\cos(\phi(x), w_y)$, the model tends to increase $||f_1||$ and decrease $||f_2||$ (i.e., activation on primitives), where $\phi(x)=a_1*f_1+a_2*f_2$ and $f_1$ is more discriminative, leading to excessive focus on the most discriminative primitives.
}
\vspace{-0.4cm}
\label{base_visual} 
\end{figure*}

Since primitives are transferred from base classes~\cite{zou2020compositional}, we then check whether the model's excessive focus on BCD primitives is inherited from biased attention during base-class training (Fig.\ref{base_visual}(a)).
To quantify this behavior, we also calculate the activation ratio on base-class samples (Fig.\ref{base_visual}(b), left), finding that the model exhibits consistently small activated regions, suggesting the collection of a limited primitive set. Further, we see that the model can still make correct predictions by using only a small number of the most attended primitives (Fig.\ref{base_visual}(b), right, \textit{pred-true}). These results confirm that the model overfits to localized, discriminative regions during base training, which we term as \textbf{regional shortcuts}. This bias persists in novel classes, limiting the potential to shift the model's attention to novel-class discriminative primitives.

\subsubsection{\textbf{Regional shortcut}}
To further delve into the cause of regional shortcuts, we analyze the optimization dynamics of representation models parameterized by nonnegative weights over primitive features, trained with cross-entropy loss. 

We consider a single sample represented by a set of primitive features $\{f_1, \dots, f_N\} \subset \mathbb{R}^d$. 
Each primitive feature $f_i$ is associated with a nonnegative weight $a_i \ge 0$, forming the weight vector $a = (a_1, \dots, a_N) \in \mathbb{R}^N$. 
The overall representation of the sample $x$ is then obtained as a weighted combination of its primitives: $\phi(x) = \sum_{i=1}^N a_i f_i$.

For a given class $y \in \{1, \dots, K\}$, let $w_y \in \mathbb{R}^d$ denote the corresponding class weight vector. 
The prediction is evaluated using a cross-entropy loss based on cosine similarity:
\begin{equation}
\setlength{\abovedisplayskip}{5pt}
\setlength{\belowdisplayskip}{5pt}
L(\phi) = - \log \frac{e^{\tau \cos(\phi, w_y)}}{e^{\tau \cos(\phi, w_y)} + \sum_{k \neq y} e^{\tau \cos(\phi, w_k)}},
\end{equation}
where $\tau$ is a temperature parameter controlling the sharpness of the distribution.

To characterize the most informative features, we define the \emph{discriminative set} of primitives $S \subseteq \{1, \dots, N\}$ as
\begin{equation}
\setlength{\abovedisplayskip}{5pt}
\setlength{\belowdisplayskip}{5pt}
S = \arg\max_{1 \leq i \leq N} \langle w_y, f_i \rangle,
\end{equation}
i.e., the primitives most aligned with $w_y$, with all maximally aligned features included in $S$.

We formalize the problem with the following assumption:
 
\textbf{Assumption 1. Non-Degradation of Competing Classes}\label{assumption1}

\textit{For any non-discriminative primitive $i \notin S$ and any discriminative primitive $s \in S$, transferring weight from $i$ to $s$ does not increase the activation or logit of any competing class $k \neq y$. In other words, the discriminative primitives in $S$ are universally superior for the ground-truth class $y$, and reallocating representation mass from outside $S$ to $S$ cannot improve the relative advantage of any alternative class. Formally, this condition can be expressed as}
\begin{equation}
\setlength{\abovedisplayskip}{5pt}
\setlength{\belowdisplayskip}{5pt}
\langle w_k - w_y, f_s - f_i \rangle \leq 0, \quad \forall i \notin S, \, s \in S, \, k \neq y,
\end{equation}
\textit{which mathematically encodes the notion that the discriminative set captures the most informative directions for the true class, ensuring that optimization dynamics favor these primitives over non-discriminative ones.}

Within the formal framework of a sample characterized by its constituent primitive features, we establish theoretical guarantees to investigate the mechanisms underlying the emergence of regional shortcuts. We initiate our analysis with these lemmas if the assumptions hold:

\textbf{Lemma 1. Gradient with Respect to Feature Weights}

For the aggregated representation $\phi = \sum_i a_i f_i$, the derivative of the loss $L$ with respect to a weight $a_i$:
\begin{equation}
\setlength{\abovedisplayskip}{5pt}
\setlength{\belowdisplayskip}{5pt}
    \frac{\partial L}{\partial a_i} 
= \sum_{k=1}^d \frac{\partial L}{\partial \phi_k} \cdot \frac{\partial \phi_k}{\partial a_i} 
= \sum_{k=1}^d (\nabla_{\phi} L)_k f_{i,k} 
= \langle \nabla_{\phi} L, f_i \rangle,
\end{equation}
Here, each component of $\phi$ contributes via the chain rule, resulting in the inner product between the gradient with respect to $\phi$ and the primitive feature $f_i$. 
Lemma~1 follows directly from this application of the chain rule, with a detailed proof provided in the Appendix.

\textbf{Lemma 2. Primitive-Level Gradient Analysis}

Consider the cross-entropy loss defined above, the gradient of the loss with respect to the aggregated representation $\phi$ is given by
\begin{equation}
\setlength{\abovedisplayskip}{5pt}
\setlength{\belowdisplayskip}{5pt}
    \nabla_{\phi} L = - w_y + \sum_{k=1}^K p_k w_k, 
\quad p_k = \frac{\exp(\langle w_k, \phi \rangle)}{\sum_{t=1}^K \exp(\langle w_t, \phi \rangle)}.
\end{equation}

Consequently, for any two primitive features $f_i$ and $f_j$ within the aggregated representation, the difference in the gradient with respect to their corresponding weights is
\begin{equation}
\setlength{\abovedisplayskip}{5pt}
\setlength{\belowdisplayskip}{5pt}
    \frac{\partial L}{\partial a_j} - \frac{\partial L}{\partial a_i} = \langle \nabla_{\phi} L, f_j - f_i \rangle = \sum_{k=1}^K p_k \langle w_k - w_y, f_j - f_i \rangle.
\end{equation}

This gradient analysis derives directly from the formal definition of the cross-entropy loss combined with the chain rule; it quantifies the contribution of each primitive feature to the aggregated representation. A detailed proof and derivation are provided in Appendix.

\textbf{Proposition 1. Gradient Comparison Between Sets} 

For any discriminative primitive $s \in S$ and any non-discriminative primitive $i \notin S$, under Assumption 1, the gradient of the loss with respect to $a_s$ is less than or equal to that with respect to $a_i$:
\begin{equation}
\setlength{\abovedisplayskip}{5pt}
\setlength{\belowdisplayskip}{5pt}
    \frac{\partial L}{\partial a_s} \leq \frac{\partial L}{\partial a_i}.
\end{equation}

This follows directly from Lemma 2, since
\begin{equation}
\setlength{\abovedisplayskip}{5pt}
\setlength{\belowdisplayskip}{5pt}
    \frac{\partial L}{\partial a_s} - \frac{\partial L}{\partial a_i}
= \sum_{k=1}^K p_k \langle w_k - w_y, f_s - f_i \rangle \leq 0,
\end{equation}
where each term is non-positive by Assumption 1 and $p_k \ge 0$. Moreover, if there exists $k$ such that $\langle w_k - w_y, f_s - f_i \rangle < 0$, the inequality is strict.  

Building upon the preceding lemmas, we now formalize the principal theoretical result, which characterizes how the model’s optimization dynamics concentrate on the discriminative set and establishes the invariant distribution criterion for feature-level representations.

\textbf{Theorem 1. Concentration on the Discriminative Set}  

Under Assumption 1, any local minimizer $a^\star$ of the loss satisfies:
\begin{equation}
\setlength{\abovedisplayskip}{5pt}
\setlength{\belowdisplayskip}{5pt}
    a^\star_j \to 0, \quad \forall j \notin S.
\end{equation}

We defer the proofs and accompanying discussion of Theorems 1 to the Appendix.
Theorem 1 has formally shown that, under Assumption 1, any local minimizer drives the weights of all non-discriminative primitives \emph{towards zero}. This conclusion follows directly from Proposition 1: since discriminative primitives consistently yield smaller gradients, transferring weight from non-discriminative to discriminative primitives never increases the loss and often strictly decreases it. As a result, the optimization dynamics naturally suppress irrelevant primitives and gradually \textbf{lead the model to concentrate on the most discriminative regions during training}.
Note that the primitive refers to a backbone-extracted local semantic unit rather than a manually defined image patch, making this primitive-level formulation generally applicable to visual backbones whose representations are formed by aggregating local primitives. Thus, the theorem should be understood as characterizing a general optimization tendency at the primitive level.

To provide intuition, we present a toy illustration of how regional shortcuts emerge during base-class training (Fig.~\ref{base_visual}(c)). Consider an image $x$ fed into a feature extractor yielding its feature map $F \in \mathbb{R}^{d \times h \times w}$, which can be divided into $h \times w$ vectors $f(a,b) \in \mathbb{R}^{d}$ corresponding to primitives in the image. For illustration, assume $h \times w=2$, resulting in $F$ containing two primitives, $f_1$ and $f_2$. we obtain the feature $\phi(x)$ corresponding to $x$ from $F$, which can be expressed simplistically as, $\phi(x)=a_1*f_1+a_2*f_2$ with weight $a_1$, $a_2$. Here, $f_1$ represents a more discriminative primitive (region), closer to the ground-truth class centroid $w_y$. To minimize the classification loss (Eq.\ref{soft}), the model shortcuts by increasing the activation of $f_1$ and decreasing that of $f_2$ to push $\phi(x)$ closer to $w_y$ (utilizing $|\cdot|_1$ to compute activations of primitives). This toy illustration thus corroborates our theoretical finding.

\vspace{-0.3cm}

\subsection{Conclusion} 
We give the following interpretation of the current model's suboptimal adaptation to novel classes: existing methods tend to minimize classification loss during the base-class training stage by exploiting highly discriminative local regions. As a result, the learned primitive set inherently lacks semantic diversity and spatial coverage. This structural bias persists when the model is transferred to novel classes, causing it to misfocus on BCD primitives during primitive composition, preventing the model's focus from shifting to novel-class discriminative regions. 
Therefore, it is essential to learn a primitive set that is semantically rich, spatially broad, and diverse while maintaining discriminability.

\section{Methodology}
\label{sec:method}
\begin{figure*}[!t]
\centering
\includegraphics[width=0.8\textwidth]{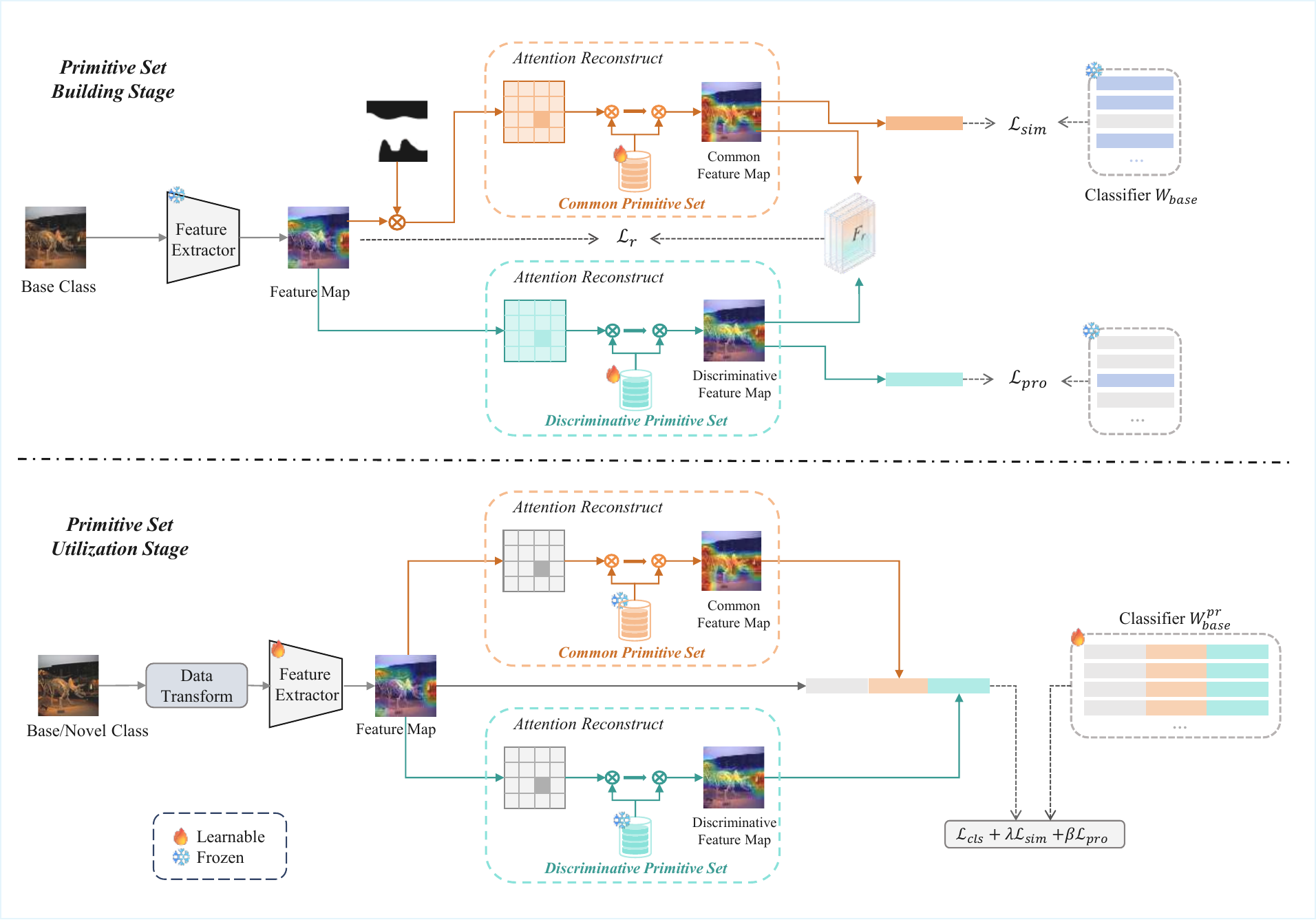} 
\vspace{-0.6cm}
\caption{ Our method comprises two main stages: the primitive set building stage and the primitive set utilization stage.
		(a) The primitive set building stage involves constructing common primitive set $R_u$ and discriminative primitive set $R_i$ based on base-class samples, with $\mathcal{L}_{sim}$, $\mathcal{L}_{pro}$, and $\mathcal{L}_{r}$ serving as constraints.
		(b) The primitive set utilization stage encourages the model to learn base and novel classes by utilizing both common and discriminative primitives. }
\label{method}
\vspace{-0.5cm}
\end{figure*}

Based on the above interpretation, we are inspired to \textit{rectify the regional shortcuts by forcing the model to learn more transferable and semantically enriched primitives, instead of over-reliance on BCD primitives for each class}. Building on this idea, we propose our method (Alleviating Regional Shortcuts by Common and Discriminative Primitives, ARS-CDP), which is divided into two stages (Fig.\ref{method}).
In the first stage, we construct a common primitive set and a discriminative primitive set based on the base classes. In the second stage, we encourage the model to learn from the common set to reduce its regional shortcuts on BCD primitives, and learn from the discriminative primitive set to facilitate learning from base classes and alleviate catastrophic forgetting.

\vspace{-0.5cm}
\subsection{Primitive Set Building Stage}

In this stage (Fig.\ref{method} top), we train the model on base classes to obtain a well-trained feature extractor $\phi(\cdot)$, a classifier $W_{base} \in \mathbb{R}^{d \times N_0}$, and two primitive sets: a common primitive set $R_{u}$ storing common primitives among base classes, and a discriminative primitive set $R_{i}$ preserving discriminative primitives between base classes. Here, $N_0$ represents the number of base classes.

During the base session, given an input image $x \in D_{train}^{0}$ labeled with $c$, the feature extractor output is $\phi(x) = g(h(x))\in\mathbb{R}^{d}$, where $h(\cdot)$ represents a encoder and $g(\cdot)$ represents a pooling layer. We obtain the feature map $F=h(x) \in \mathbb{R}^{d \times\!h\times\!w}$ from the encoder. Each patch \( f(a,b) \in \mathbb{R}^d \) in \( F \) corresponds to a distinct spatial location, representing candidate primitives \cite{zou2024compositional}. The primitive granularity is determined by the output resolution of the backbone, where each spatial/token location in the final backbone representation is treated as one primitive. Accordingly, the image \( x \) can be represented as the primitive set
$\{ f(a,b) \mid a \in \{0, \ldots, h-1\},\ b \in \{0, \ldots, w-1\} \}$. From this set, we generalize the common primitive set and the discriminative primitive set.

Previous methods such as Class Activation Mapping (CAM)\cite{zhou2016learning} rely on implicit constraints to suppress over-reliance on biased local regions, but lack explicit structural constraints to ensure semantic consistency and knowledge sharing across classes. Consequently, their ability to promote transferable representations in few-shot class-incremental learning is limited. In contrast, our method employs explicit constraints by decomposing features into common and discriminative primitive sets, facilitating semantic sharing and class-specific discrimination. This explicit modeling effectively alleviates regional shortcuts and better supports knowledge transfer and forgetting mitigation in incremental learning.

\textbf{Common Primitive Set:} To mitigate the model's regional shortcuts on discriminative primitives during the base session, we encourage the model to represent images using common primitives. Drawing inspiration from the Basic Level Theory \cite{rosch1976basic} in cognitive science, which posits that basic-level categories offer higher cognitive efficiency and richer semantic content, we design a common primitive set to comprise $M$ adaptive primitive sub-sets. Each sub-set corresponds to a basic-level category and is denoted as $R_{u} = \left\{r_u^1,\ldots,r_u^M\right\}$, where $r_u^M \in \mathbb{R}^{d \times U}$. Every adaptive primitive subset $r_{u}^m = \left\{u_1^m,\ldots,u_U^m\right\}$ contains a set of common primitives for a specific basic-level category, with $u_U^m \in \mathbb{R}^{d}$. These sub-sets are not manually assigned to predefined semantic categories; instead, they emerge as latent semantic structures learned from the training data. To maintain feature sparsity within each sub-repository, we adopt orthogonal initialization for each sub-primitive and impose an orthogonal loss during training, as proposed in \cite{liu2023learning}.

To learn the common primitive set, we explore the common primitives to capture inter-class relationships. Therefore, we first identify primitives through a common-feature mask, then highlight the inter-class relationships by removing the prediction of each sample's ground truth class, and finally learn the common primitive set based on the reconstructed primitives and the class similarities.
Specifically, we first compute a patch-wise commonness score for each spatial primitive. This score measures its association with semantically related classes and indicates whether the primitive tends to capture shared information rather than class-specific patterns. We then feed the score into a two-layer MLP to generate a spatial mask $M_u \in \mathbb{R}^{1 \times h \times w}$ and obtain the common feature map $F_u = M_u \odot F$, where $F \in \mathbb{R}^{d \times h \times w}$ denotes the original feature map, since it provides a lightweight learnable mapping that is more flexible than other methods such as thresholding or attention pooling.

Next, we reconstruct the common feature map $F_u=\{{f}_{u}(a,b)\}_{a<h,b<w}$ into $\bar{F}_u = \{\bar{f}_{u}(a,b)\}_{a<h,b<w}$ using an attention mechanism:
\begin{equation}\label{atten1}
\setlength{\abovedisplayskip}{5pt}
\setlength{\belowdisplayskip}{5pt}
\begin{array}{c}
\displaystyle \bar{f}_{u}(a,b) = \sum_{i=1}^{M}\sum_{j=1}^{U} e_{ij} u_j^i \\[8pt]
\displaystyle e_{ij} = \frac{\exp(f_u(a,b) u_j^i)}{\sum_{i=1}^{M} \sum_{j=1}^{U} \exp(f_u(a,b) u_j^i)},
\end{array}
\end{equation}
Based on $\bar{F}_u$, we propose a similarity-guided loss that \textbf{encourages the common feature map to capture inter-base-class similarity}, ensuring the transferability of common primitives within $R_u$ across base classes. First, to capture inter-base-class similarities, we compute the similarity between the image $x$ and all base classes to obtain the label-similar class:
\begin{equation}
\setlength{\abovedisplayskip}{5pt}
\setlength{\belowdisplayskip}{5pt}
	\label{simi}I(x)= \quad (\frac{W_{base}}{\left\|W_{base}\right \|_{2} })^T (\frac{w_c}{\left \|w_c\right\|_{2}}),
\end{equation}
where $w_c \in \mathbb{R}^{d}$ represents the classifier for ground-truth label. 
The output $I(x) \in \mathbb{R}^{N_0}$ indicates the similarity between the image $x$ and all base classes. Next, we calculate the probability $Pb(\bar{F}u) \in \mathbb{R}^{N_0}$ that the common feature map $\bar{F}_u$ is categorized into each base class: 
\begin{equation}\label{proba}
\setlength{\abovedisplayskip}{5pt}
\setlength{\belowdisplayskip}{5pt}
		Pb(\bar{F}_u)= \quad (\frac{W_{base}}{\left\|W_{base}\right \|_{2} })^T (\frac{g(\bar{F}_u)}{\left \|g(\bar{F}_u)\right\|_{2}}),
\end{equation}
Finally, the similarity-guided loss is expressed as:
\vspace{-0.1cm}
\begin{equation}\label{simi_loss}
\setlength{\abovedisplayskip}{5pt}
\setlength{\belowdisplayskip}{5pt}
		\mathcal{L}_{sim}= \text{JS}(o(Pb(\bar{F}_u)) \parallel o(I(x))),
\end{equation}
where $\text{JS}(\parallel)$ represents Jensen-Shannon Divergence \cite{hjelm2018learning}, a measure often used to quantify the difference between a predicted probability distribution and true labels.

To ensure that the common feature map captures class-agnostic shared semantics rather than class-specific discriminative cues, we apply the operator $o(\cdot)$ to suppress the contribution of the ground-truth class by setting its corresponding value to zero, thereby preventing trivial alignment from label supervision and preserving of inter-class relational structure essential for learning transferable common primitives.

\textbf{Discriminative Primitive Set:} To maintain the model's performance on base classes, we propose to construct a discriminative primitive set $R_{i}$ alongside $R_{u}$. This set consists of a set of discriminative primitives, denoted as $R_{i} = \left\{u_1,\ldots,u_I\right\}$,  where $u_I \in \mathbb{R}^{d}$ represents a discriminative primitive. To enable the discriminative primitive set to adaptively store discriminative features, we introduce a prototype-centered loss. First, we reconstruct the original feature map $F$ into a discriminative feature map $F_i$ using the discriminative primitive set with an attention mechanism, following Eq.\ref{atten1} (where the common primitive set is replaced by the discriminative set).

To ensure the discriminative primitive set focuses on class-specific features, we design a loss that \textbf{forces the discriminative feature map to emphasize primitives correctly associated with the ground-truth class}. This design indirectly encourages elements in the discriminative primitive set to capture discriminative features in base classes.

Specifically, based on the reconstructed discriminative feature map $F_i \in \mathbb{R}^{d \times h\times w}$, we apply a prototype-centered loss to ensure correct categorization into the ground-truth class, thus enhancing the discriminative features in the set. The prototype-centered loss is defined as:
\begin{equation}\label{proto_loss}
\setlength{\abovedisplayskip}{5pt}
\setlength{\belowdisplayskip}{5pt}
		\mathcal{L}_{pro}= 1-cos((\frac{g(F_i)}{\left\| g(F_i)\right\|_{2}})^T,\frac{w_c}{\left\| w_c\right\|_{2}}).
\end{equation}
Here, \(w_c\) denotes the classifier weight of the ground-truth class \(c\). It is initialized as part of the base classifier and adaptively optimized during base-class training, thus serving as an adaptive class prototype.

To jointly train the two sets and encourage their transferable and class-discriminative focuses, we introduce a primitive-reconstruction loss:
{\small
\begin{equation}\label{recon1}
\setlength{\abovedisplayskip}{5pt}
\setlength{\belowdisplayskip}{5pt}
F_{r} = (1 - Ratio_i) \bar{F}_u + Ratio_i F_i, \quad
Ratio_i = \frac{I}{M \times U + I}
\end{equation}
}
\vspace{-0.4cm}
{\small
\begin{equation}\label{recon2}
\setlength{\abovedisplayskip}{5pt}
\setlength{\belowdisplayskip}{5pt}
\mathcal{L}_{r} = \text{CE} \Big(
\tau \cdot \cos \big(
\left(\frac{F_{r}}{\| F_{r} \|_2}\right)^T\!,\frac{F}{\| F \|_2}\big), 
 \text{diag}\big(0,1,\ldots,(h \times w - 1)\big)
\Big),
\end{equation}
}
where $F_r \in \mathbb{R}^{d \times h \times w}$ denotes the reconstructed feature map obtained by combining $F_i$ and $\bar{F}_u$. In this context, $M$ is the number of adaptive sub-sets in the common primitive set $R_u$, $U$ is the number of primitives within each sub-set, and $I$ is the number of primitives in the discriminative primitive set $R_i$. Consequently, $M \times U$ represents the total number of common primitives, and $M \times U + I$ denotes the total number of primitives across both sets. The ratio $\text{Ratio}_i = \frac{I}{M \times U + I}$ determines the relative weighting of discriminative versus common primitives during reconstruction, thereby balancing generalizable and class-specific information.  

\renewcommand{\arraystretch}{1.1} 
\begin{table*}[!t]
\caption{Top-1 average accuracy on all seen classes. Results for the compared methods are sourced from \protect\cite{yang2023neural}.
\label{tab: result_cifar}}
\vspace{-0.2cm}
\centering
\adjustbox{max width=\textwidth}{
\begin{tabular}{l l ccccccccc c}
\toprule
\multirow{2}{*}{\textbf{Method}} & \multirow{2}{*}{\textbf{Backbone}} & \multicolumn{9}{c}{\textbf{Accuracy in each session(\%)} on CIFAR-100} & \multirow{2}{*}{\textbf{Average}} \\
                                 &                                    & \textbf{0} & \textbf{1} & \textbf{2} & \textbf{3} & \textbf{4} & \textbf{5} & \textbf{6} & \textbf{7} & \textbf{8} & ~ \\
\cmidrule(lr){3-11}
CEC\cite{zhang2021few}             & ResNet-12      & 73.07      & 68.88      & 65.26      & 61.19      & 58.09      & 55.57      & 53.22      & 51.34      & 49.14      & 59.53 \\
LIMIT\cite{zhou2022few}           & ResNet-12      & 73.81      & 72.09      & 67.87      & 63.89      & 60.70      & 57.77      & 55.67      & 53.52      & 51.23      & 61.84 \\
Meta FSCIL\cite{chi2022metafscil}      & ResNet-12      & 74.50      & 70.10      & 66.84      & 62.77      & 59.48      & 56.52      & 54.36      & 52.56      & 49.97      & 60.79 \\
Self-promoted\cite{zhu2021self}   & ResNet-12      & 64.10      & 65.86      & 61.36      & 57.45      & 53.69      & 50.75      & 48.58      & 45.66      & 43.25      & 54.52 \\
FACT\cite{zhou2022forward}            & ResNet-12      & 74.60      & 72.09      & 67.56      & 63.52      & 61.38      & 58.36      & 56.28      & 54.24      & 52.10      & 62.24 \\
Data-free Replay\cite{liu2022few} & ResNet-12      & 74.40      & 70.20      & 66.54      & 62.51      & 59.71      & 56.58      & 54.52      & 52.39      & 50.14      & 60.78 \\
ALICE\cite{peng2022few}           & ResNet-12      & 79.00      & 70.50      & 67.10      & 63.40      & 61.20      & 59.20      & 58.10      & 56.30      & 54.10      & 63.21 \\
NC-FSCIL\cite{yang2023neural}        & ResNet-12      & 82.52      & 76.82      & 73.34      & 69.68      & 66.19      & 62.85      & 60.96      & 59.02      & 56.11      & 67.50 \\
RDI\cite{zhou2024delve}             & ResNet-12      & 81.45      & 77.02      & 72.73      & 68.95      & 65.75      & 63.02      & 61.07      & 59.01      & 56.72      & 67.30 \\
R-FSCIL\cite{tang2024rethinking}         & ResNet-12      & \textbf{82.90} & 76.30 & 72.90 & 67.80 & 65.20 & 62.00 & 60.70 & 58.80 & 56.60 & 67.02 \\
\textbf{Ours}   & ResNet-12      & 81.25      & \textbf{77.26} & \textbf{73.56} & \textbf{69.60} & \textbf{66.64} & \textbf{63.84} & \textbf{62.14} & \textbf{60.62} & \textbf{58.28} & \textbf{68.13} \\
PriViLege\cite{park2024pre}       & ViT-B/16       & 90.88      & 89.39      & 88.97      & 87.55      & 87.83      & 87.35      & 87.53      & 87.15      & 86.06      & 88.08 \\
\textbf{Ours}   & ViT-B/16       & \textbf{92.68} & \textbf{90.37} & \textbf{89.47} & \textbf{88.29} & \textbf{88.25} & \textbf{87.82} & \textbf{87.79} & \textbf{87.33} & \textbf{86.19} & \textbf{88.69} \\
\bottomrule
\end{tabular}}
\vspace{-0.5cm}
\end{table*}
\renewcommand{\arraystretch}{1.0} 

The cosine similarity term $\text{cos}\!\left( \left(\frac{F_r}{\left\| F_r \right\|_2} \right)^{\!T}, \frac{F}{\left\| F \right\|_2} \right) \in \mathbb{R}^{h \times w}$ measures similarity between the reconstructed and original feature maps along the channel dimension $c$, while $\text{diag}(\cdot)$ denotes the label matrix used in the loss computation. By minimizing this loss, the reconstructed feature map $F_r$ is encouraged to closely match the original feature map $F$. This optimization guides the common primitive set to focus on more generalizable, transferable patterns, while the discriminative primitive set concentrates on class-specific, highly discriminative cues, enabling a complementary division of roles for more effective sample characterization.

\vspace{-0.3cm}

\subsection{Primitive Set Utilization Stage}
During the primitive set utilization stage (the bottom of Fig.\ref{method}), we fix the common primitive set and discriminative primitive set and compel the model to utilize them for base-class classification, resulting in a new feature extractor $\phi^{pr}(\cdot)$ and a new classifier $W^{pr}_{base}$. Fixing these sets is a reasonable strategy commonly adopted in incremental learning \cite{ho2023prototype, wang2023few}, as base classes have sufficient data to learn robust primitives, and freezing them helps prevent overfitting to scarce novel-class samples and mitigates catastrophic forgetting. 
Meanwhile, the feature extractor $\phi^{pr}(\cdot)$ is fine-tuned with a small learning rate throughout this stage to adaptively refine representations for both base and novel classes, allowing for novel knowledge outside the primitive sets. 
To expedite model convergence, we initialize $W^{pr}_{base}$ with the pre-trained feature extractor $W_{base}$ from the first stage.

To train the model using the common primitive set and discriminative primitive set to represent base classes and prevent it from being trapped in local optima due to initialization, we perform mixed-sample transformations on base-class samples, inspired by \cite{yang2023neural}. 
Specifically, we combine two images in each mini-batch using a random binary spatial mask and mix their labels according to the mask ratio, encouraging more transferable primitive combinations and reducing overfitting to fixed primitive regions.
Subsequently, we generate the original feature map $F^{pr}$ for base-class samples based on $W^{pr}_{base}$ and reconstruct $F^{pr}$ into common feature map $F^{pr}_u$ and discriminative feature map $F^{pr}_i$ using the common primitive set $R_u$ and discriminative primitive set $R_i$.
Finally, we apply global average pooling to the original, common, and discriminative feature maps separately, and concatenate the resulting embeddings along the channel dimension to obtain the primitive-augmented image representation \(\phi^{pr}(x) \in \mathbb{R}^{3c}\). After normalization, \(\phi^{pr}(x)\) is fed into a dimension-matched cosine classifier, where the logits are computed by cosine similarity between the concatenated embedding and each class prototype, followed by the cross-entropy loss.

During this stage, we continue to use $\mathcal{L}_{sim}$ and $\mathcal{L}_{pro}$ to govern the properties of the two sets. The final loss function can be expressed as:
\begin{equation}
\setlength{\abovedisplayskip}{5pt}
\setlength{\belowdisplayskip}{5pt}
    \mathcal{L}=\mathcal{L}_{ce}+\lambda \mathcal{L}_{sim}+\beta \mathcal{L}_{pro},
\end{equation}
where $\lambda$ and $\beta$ are hyperparameters.

Moving to the incremental session, each base class's classifier is replaced with the class's average embedding. The prototype calculation is denoted as:
\begin{equation}
\setlength{\abovedisplayskip}{5pt}
\setlength{\belowdisplayskip}{5pt}
    p_c = \frac{1}{n^0_c} \textstyle \sum_{j=1}^{D_{train}^0} \mathbb{I}(y_j=c) \phi^{pr} (x_j),
\end{equation}
where $\mathbb{I}$ represents the indicator function and $n^0_c$ denotes the count of examples in class $c$.

Then the classifier for base classes remains fixed, while for novel classes, the classifier can be refined by employing the average embedding computed in the primitive-augmented representation space. The learned common and discriminative primitive sets are reused during novel-class adaptation, avoiding the reconstruction of new primitive sets from scarce novel samples. During evaluation in session $T$, an input $x$ is predicted using its primitive-augmented feature $\phi (x)^{pr}$ and the classifier $W^{pr}$, as stated in Sec.~\ref{protonet}.

\vspace{-0.1cm}
\section{Experiment}
\label{sec:experiments}
\vspace{-0.1cm}
\subsection{Dataset and Implementation Details}
\noindent\textbf{Dataset \& Split:} 
Following \cite{yang2023neural}, we validate our method on CIFAR-100 \cite{krizhevsky2009learning}, \textit{mini}ImageNet \cite{russakovsky2015imagenet}, and CUB-200-2011 (CUB200) \cite{wah2011caltech}, using consistent data splits. CIFAR-100 contains 60,000 low-resolution images across 100 classes. miniImageNet, derived from ImageNet, includes 100 classes with 600 medium-resolution images each, commonly used for few-shot and incremental learning. CUB-200 comprises 11,788 images of 200 bird species, offering a fine-grained and challenging benchmark for recognition and transfer learning.

\noindent\textbf{Training Details:} 
Following\cite{yang2023neural}, we use ResNet-12 and ImageNet-pretrained ViT-B/16 for CIFAR-100, ResNet-12 for \textit{mini}ImageNet, and ImageNet-pretrained ResNet-18 for CUB200.
Experiments using ImageNet-pretrained ViT-B/16 and ResNet-18 are conducted on NVIDIA A5000 and NVIDIA GeForce RTX 3090 GPUs, while experiments using ResNet-12 are conducted on NVIDIA GeForce GTX 1080 GPUs. The backbone choices follow standard FSCIL evaluation settings for fair comparison, and our method is backbone-agnostic.

\vspace{-0.4cm}
\subsection{Comparison with State-of-the-Arts}

\begin{figure}[t]
\centering
\includegraphics[width=0.8\linewidth]{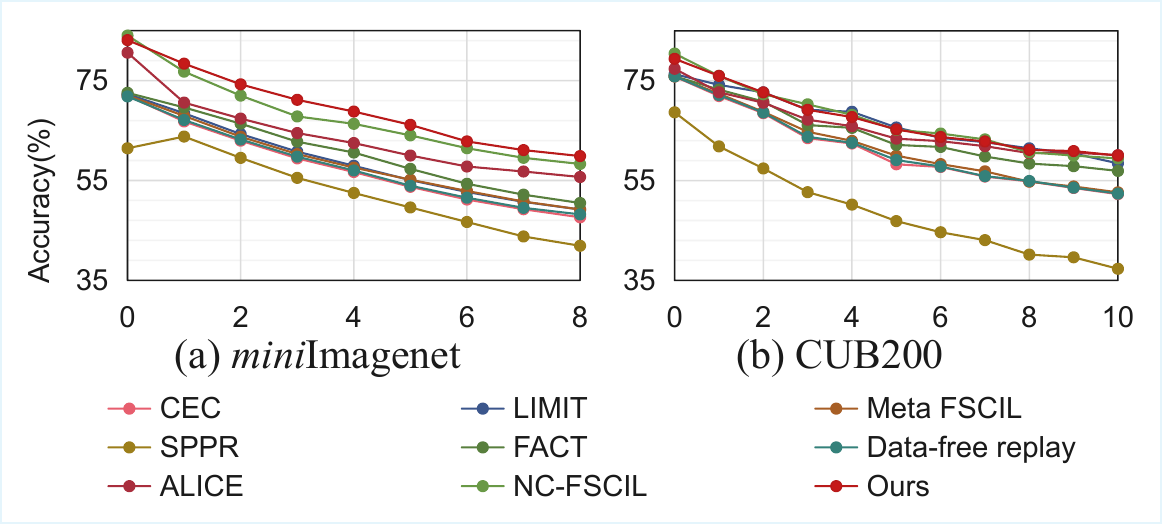}
\vspace{-0.4cm}
\caption{
   Comparison during each incremental session.
}
\label{fig: result} 
\vspace{-0.2cm}
\end{figure}

Across three widely used benchmarks, our method consistently outperforms state-of-the-art approaches. On CIFAR-100 (Tab.~\ref{tab: result_cifar}), it surpasses CEC~\cite{zhang2021few}, LIMIT~\cite{zhou2022few}, Meta FSCIL~\cite{chi2022metafscil}, and recent methods like NC-FSCIL~\cite{yang2023neural}, RDI~\cite{zhou2024delve}, and R-FSCIL~\cite{tang2024rethinking}, achieving 68.13\% top-1 accuracy with ResNet-12 and 88.69\% with ViT-B/16, exceeding PriViLege~\cite{park2024pre}. Fig.~\ref{fig: result} shows consistent superior performance across incremental sessions on miniImageNet and CUB-200, demonstrating robustness and effectiveness. For fair comparison, all methods with the same backbone follow the same training protocol. Additional experimental results and analysis are provided in the appendix to further demonstrate our approach’s superiority and generalizability.

\vspace{-0.4cm}
\subsection{Ablation Study}

We conduct ablation studies on CIFAR-100 to investigate the individual and combined contributions of the common and discriminative primitive sets (Tab.\ref{tab: ablation_study}). 
The results indicate that using these primitive sets separately already enhances the model’s generalization. Specifically, the common set improves generalization to novel classes, crucial for incremental learning with limited new samples, while the discriminative set mainly helps maintain and even boost performance on base classes, forming a solid foundation for learning.
\begin{table}[!t]
\centering
\caption{Ablation study of each module.
}
\vspace{-0.2cm}
\label{tab: ablation_study}
\adjustbox{max width=\columnwidth}{
\begin{tabular}{ccccccc}
\toprule
$\mathcal{L}_{sim}$ & $\mathcal{L}_{pro}$ & $\mathcal{L}_{r}$ & novel & base & average\_acc & harmonic\_acc \\
\midrule
& & & 18.63 & 81.22 & 54.28 & 48.34 \\
\checkmark & & & 22.45 & 81.50 & 56.29 & 50.65 \\
\checkmark & \checkmark & & 23.40 & 81.98 & 56.72 & 51.17 \\
\checkmark & \checkmark & \checkmark & \textbf{29.97} & \textbf{82.03} & \textbf{58.28} & \textbf{53.56} \\
\bottomrule
\end{tabular}
}
\vspace{-0.3cm}
\end{table}

\begin{figure}[!t]
\centering
\includegraphics[width=0.9\linewidth]{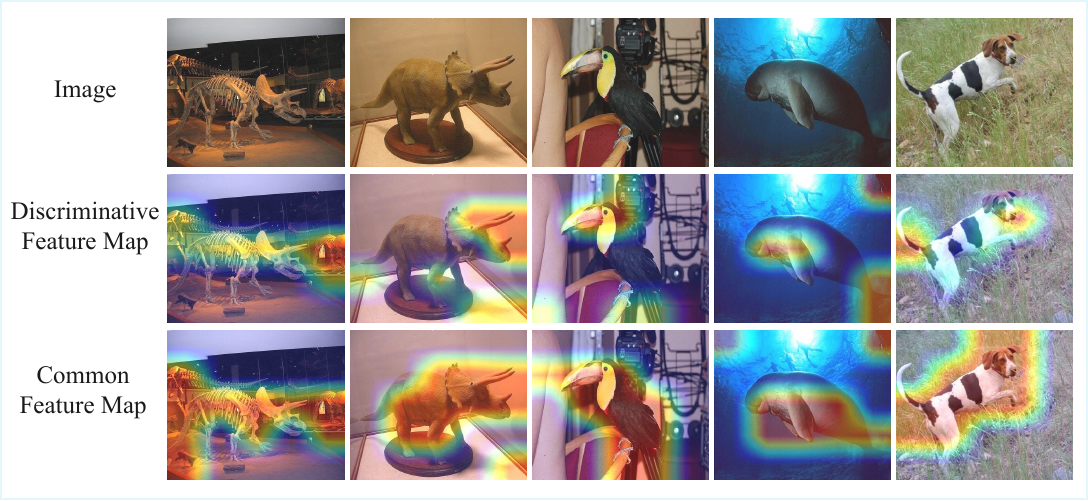} 
\vspace{-0.3cm}
\caption{
   Visualization shows that the common feature map $F^{pr}_u$ tends to focus on the entire regions relevant to the class. Although not all regions are discriminative, the focus on less-discriminative regions benefits the transferability across novel and base classes. In contrast, the discriminative feature map $F^{pr}_i$ tends to focus on the local region with discriminability.
}
\label{fig: primitive_map} 
\vspace{-0.2cm}
\end{figure}

To further illustrate the distinct roles and complementary nature of these two primitive sets, we visualize the feature maps reconstructed by each set in Fig.\ref{fig: primitive_map}. 
The discriminative feature map $F^{pr}_i$ focuses on localized, class-specific regions with strong discriminative power for distinguishing between categories. 
In contrast, the common feature map $F_{pr}^u$ highlights broader regions relevant to the class, capturing shared and transferable semantic information.  

\vspace{-0.3cm}

\begin{figure}[H]
\centering
\includegraphics[width=0.7\linewidth]{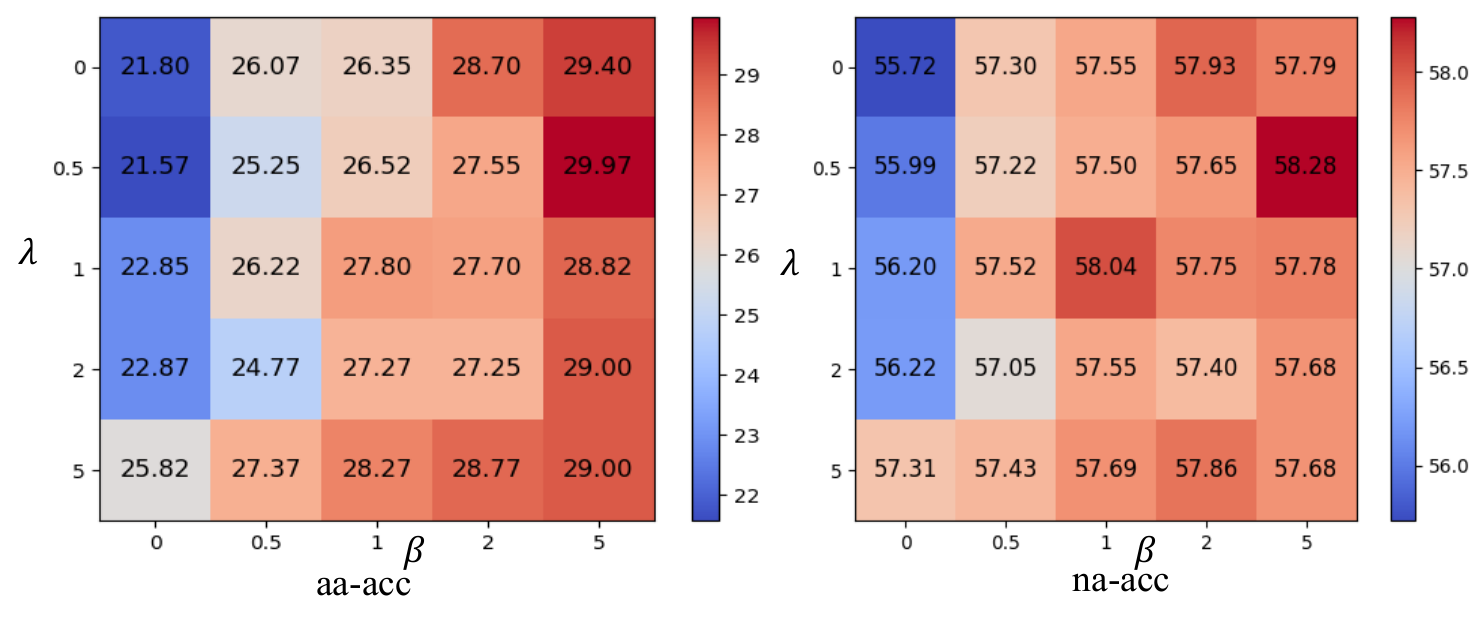} 
\vspace{-0.4cm}
\caption{
   Influence of hyper-parameters rules out data augmentation as the cause of performance improvement, indirectly validating our method design.
}
\vspace{-0.3cm}
\label{fig: parameter} 
\end{figure}

Additionally, we conduct hyper-parameter sensitivity experiments on CIFAR-100 (Fig.\ref{fig: parameter}) to evaluate the impact of the second-stage loss components on two metrics: 'aa-acc' (overall accuracy) and 'na-acc' (novel-class accuracy). When both hyperparameters \(\lambda\) and \(\beta\) are set to zero—removing the similarity loss \(L_{sim}\) and prototype loss \(L_{pro}\)—performance drops significantly on both metrics. This demonstrates that the observed improvements are driven by the designed loss functions enforcing semantic consistency and discriminability, rather than trivial factors such as data augmentation, confirming the effectiveness and robustness of our method.

\begin{table}[H]
\centering
\vspace{-0.2cm}
\caption{Ablation study for parameters M, U, and I on CIFAR-100.}
\label{tab:ablation_parameters}
\begin{tabular}{lccccccc}
\toprule 
M & 2 & 6 & 4 & 4 & 4 & 4 & 4 \\
U & 60 & 60 & 40 & 80 & 60 & 60 & 60 \\
I & 240 & 240 & 240 & 240 & 200 & 280 & 240 \\
\midrule
Acc. & 56.18 & 57.35 & 56.96 & 57.73 & 57.49 & 57.37 & 58.28 \\
\bottomrule
\end{tabular}
\vspace{-0.2cm}
\end{table}

We conduct experiments on CIFAR-100 to verify the impact of parameters M, U, and I, and the results are illustrated as Tab.\ref{tab:ablation_parameters}. On \textit{mini}ImageNet, we set M=3, U=100, and I=300. On CUB200, we set M=2, U=100, and I=200.

\subsection{Qualitative and Quantitative Analysis }

\begin{figure}[!t]
\centering
\includegraphics[width=\linewidth]{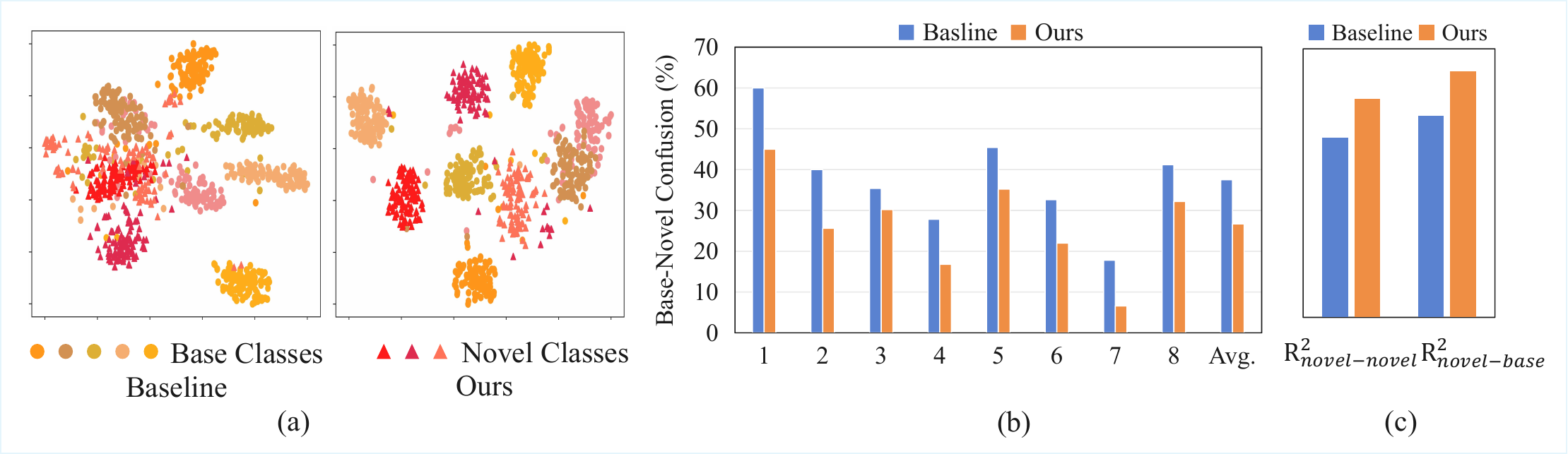} 
\vspace{-0.7cm}
\caption{
   (a) Comparison of distributions reveals that our method enhances the compactness of base-class and novel-class clusters and mitigates misclassification.
   (b) Comparison of the misclassification shows that ours mitigates it in each session. 
   (c) Comparison of class separation indicates that ours improves novel-novel class and base-novel class separation.
}
\label{fig: distribution} 
\vspace{-0.2cm}
\end{figure}

\textbf{Alleviating Feature-Space Confusion:} 
To validate our method in mitigating confusion between novel and base classes, we use t-SNE \cite{van2008visualizing} to visualize and compare feature embeddings from the baseline and our approach. We randomly select 5 base classes (circles) and 3 novel classes (triangles) from CIFAR-100, including all test samples for visualization.
As shown in Fig.\ref{fig: distribution}(a), the baseline exhibits overlap between novel and base features, leading to frequent misclassification. In contrast, our method produces more compact and well-separated clusters, reducing confusion and improving class discriminability. This clearer separation validates the model’s ability to distinguish novel from base classes, enhancing performance and robustness in incremental learning.

\textbf{Mitigating Novel-Base Classification Confusion:}
We quantitatively evaluate our method on CIFAR-100 by measuring its ability to reduce the misclassification of novel classes as base classes across incremental sessions. Specifically, we compare the accuracy of novel samples over novel classes only (sn-acc) with their accuracy over all encountered classes (sa-acc), where a larger gap indicates greater confusion. As shown in Fig.\ref{fig: distribution}(b), our method consistently narrows this gap throughout incremental learning, demonstrating improved separation between novel and base classes. Despite session-wise variations due to class distribution and difficulty, the overall trend confirms enhanced classification clarity and robustness.

\textbf{Improving Separation:} 
Following~\cite{song2023learning}, we compute the separation degree $R^2$ to quantify the distinction between base and novel classes in the learned feature space, where larger values indicate better separation. As shown in Fig.\ref{fig: distribution}(c) on miniImageNet, our method achieves higher $R^2_{\text{novel-base}}$ and $R^2_{\text{novel-novel}}$ than the baseline, demonstrating improved separation both between base and novel classes and among novel classes themselves. This indicates that the learned primitive sets are transferable and semantically enriched, enabling effective recomposition while enhancing fine-grained discriminability in incremental learning.
\vspace{-0.4cm}
\begin{figure}[H]
\centering
\includegraphics[width=\linewidth]{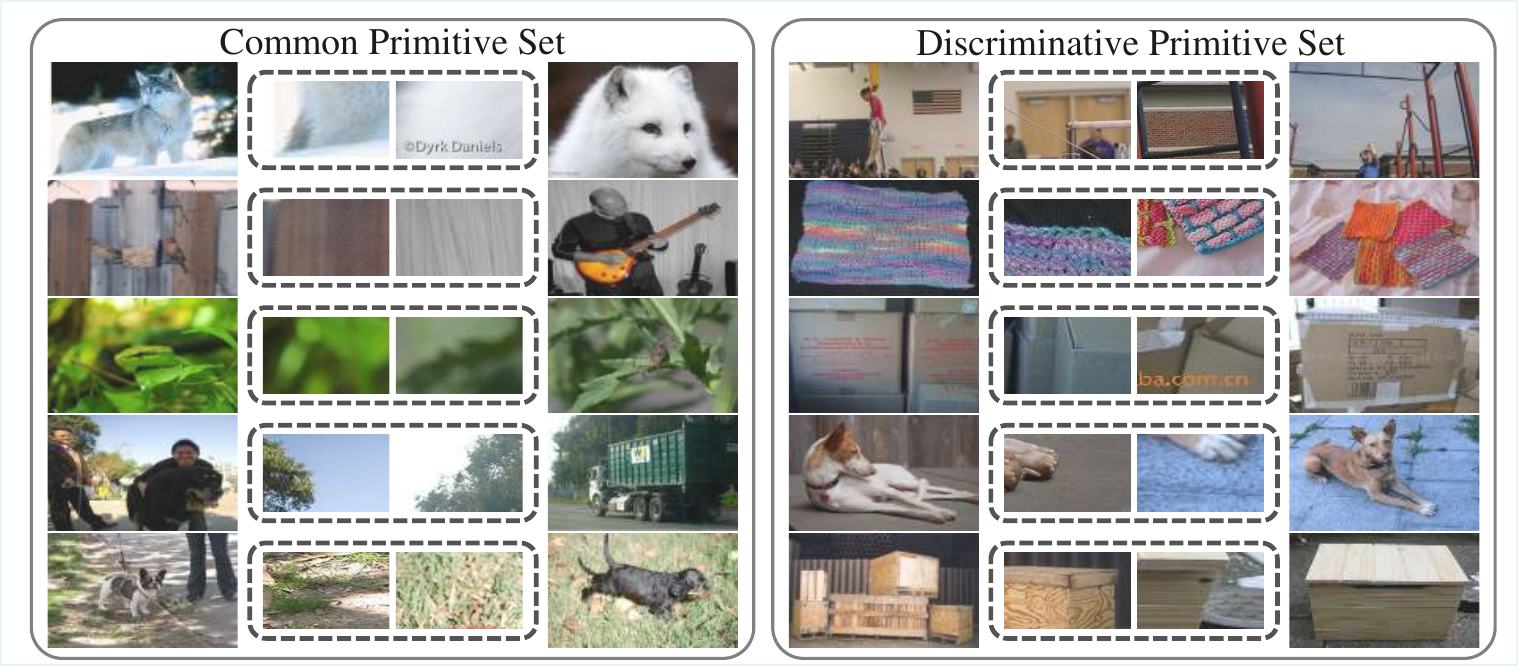} 
\vspace{-0.7cm}
\caption{
  The common primitive set captues inter-class common features, whereas the discriminative set captures intra-class discriminative features, demonstrating the interpretability of our method for the model.
}
\label{fig: set} 
\vspace{-0.2cm}
\end{figure}

\textbf{Interpretability of Primitive Sets:} 
To understand the semantic meaning of the two primitive sets, we retrieve image patches from \textit{mini}ImageNet most similar to each primitive. As shown in Fig.~\ref{fig: set}, the common set captures cross-class features like shared textures, colors, or backgrounds (e.g., white fur in Arctic wolves and malamutes, green-leaf backgrounds in harvestmen and grass snakes), facilitating semantic transfer. The discriminative set encodes fine-grained, within-class features (e.g., Ibizan Hound’s feet, carton edges), crucial for distinguishing similar classes. Together, they provide complementary representations, enabling generalization to unseen classes while maintaining discrimination among base classes.

\begin{table}[H]
\centering
\vspace{-0.2cm}
\caption{Comparison of decision margins. A larger margin indicates better separation between the ground-truth class and its strongest competitor.}
\label{tab:decision_margin}
\resizebox{0.78\columnwidth}{!}{
\begin{tabular}{lccc}
\toprule
Methods & Base Margin & Novel Margin & All Margin \\
\midrule
Baseline & 0.3013 & 0.2052 & 0.2629 \\
\textbf{Ours} & \textbf{0.3328} & \textbf{0.2067} & \textbf{0.2663} \\
\bottomrule
\end{tabular}
}
\vspace{-0.2cm}
\end{table}

\textbf{Verification of Assumption 1:}
To empirically validate Assumption 1 in real FSCIL scenarios, we measure the logit margin between the ground-truth class and its strongest competing class, i.e., $z_y - \max_{k \neq y} z_k$. As shown in Tab.\ref{tab:decision_margin}, our method consistently yields larger margins than the baseline, indicating better separation between the target class and competing classes. This suggests that the proposed primitive decomposition indeed improves discriminative separation in practical FSCIL settings.

\begin{table}[H]
\centering
\vspace{-0.2cm}
\caption{Comparison under different mask generation strategies on CIFAR-100 with ViT-B/16.}
\label{tab:mask_strategy}
\resizebox{\columnwidth}{!}{
\begin{tabular}{lcccccccccc}
\toprule
Method & \multicolumn{9}{c}{Accuracy in each session (\%) on CIFAR-100} & Avg. \\
\cmidrule(lr){2-10}
 & 0 & 1 & 2 & 3 & 4 & 5 & 6 & 7 & 8 &  \\
\midrule
Baseline & 90.43 & 88.08 & 87.69 & 86.56 & 86.65 & 86.41 & 86.37 & 86.08 & 84.83 & 87.01 \\
Attention Pooling & 91.35 & 88.89 & 88.34 & 87.19 & 87.17 & 86.75 & 86.71 & 86.41 & 85.28 & 87.57 \\
Thresholding & 90.35 & 87.85 & 87.51 & 86.36 & 86.46 & 86.20 & 86.21 & 85.97 & 84.78 & 86.85 \\
\textbf{Ours} & \textbf{92.68} & \textbf{90.37} & \textbf{89.47} & \textbf{88.29} & \textbf{88.25} & \textbf{87.82} & \textbf{87.79} & \textbf{87.33} & \textbf{86.19} & \textbf{88.69} \\
\bottomrule
\end{tabular}
}
\vspace{-0.2cm}
\end{table}

\textbf{Mask Generation Strategy:}
To evaluate the design of our mask generation module, we compare the proposed MLP-based strategy with two simpler alternatives: attention pooling and thresholding. As shown in Table~\ref{tab:mask_strategy}, the MLP-based strategy achieves the best average accuracy, demonstrating the benefit of learnable mask calibration for common primitive extraction.

\begin{table}[H]
\centering
\vspace{-0.2cm}
\caption{Ablation study of primitive sets on CIFAR-100 with ViT-B/16.}
\label{tab:primitive_set_ablation}
\resizebox{\columnwidth}{!}{
\begin{tabular}{lcccccccccc}
\toprule
Method & 0 & 1 & 2 & 3 & 4 & 5 & 6 & 7 & 8 & Avg. \\
\midrule
Baseline & 90.43 & 88.08 & 87.69 & 86.56 & 86.65 & 86.41 & 86.37 & 86.08 & 84.83 & 87.01 \\
Only Common & 91.27 & 88.88 & 88.36 & 87.25 & 87.31 & 86.96 & 86.92 & 86.55 & 85.38 & 87.65 \\
Only Discriminative & 91.30 & 88.80 & 88.37 & 87.31 & 87.30 & 86.89 & 86.83 & 86.45 & 85.28 & 87.61 \\
\textbf{Ours} & \textbf{92.68} & \textbf{90.37} & \textbf{89.47} & \textbf{88.29} & \textbf{88.25} & \textbf{87.82} & \textbf{87.79} & \textbf{87.33} & \textbf{86.19} & \textbf{88.69} \\
\bottomrule
\end{tabular}
}
\vspace{-0.2cm}
\end{table}

\textbf{Ablation on Primitive Sets:}
We evaluate the contribution of the common and discriminative primitive sets. As shown in Table~\ref{tab:primitive_set_ablation}, both the common-only and discriminative-only variants improve over the baseline, while their combination achieves the best performance. This demonstrates that the two primitive sets provide complementary benefits.

\begin{table}[H]
\centering
\vspace{-0.2cm}
\caption{Comparison with shortcut-reducing data augmentation on CIFAR-100 with ViT-B/16.}
\label{tab:shortcut_aug}
\resizebox{\columnwidth}{!}{
\begin{tabular}{lcccccccccc}
\toprule
Method & 0 & 1 & 2 & 3 & 4 & 5 & 6 & 7 & 8 & Avg. \\
\midrule
Baseline & 90.43 & 88.08 & 87.69 & 86.56 & 86.65 & 86.41 & 86.37 & 86.08 & 84.83 & 87.01 \\
Data Augmentation & 90.52 & 88.22 & 87.77 & 86.64 & 86.69 & 86.67 & 86.45 & 86.26 & 84.95 & 87.13 \\
\textbf{Ours} & \textbf{92.68} & \textbf{90.37} & \textbf{89.47} & \textbf{88.29} & \textbf{88.25} & \textbf{87.82} & \textbf{87.79} & \textbf{87.33} & \textbf{86.19} & \textbf{88.69} \\
\bottomrule
\end{tabular}
}
\vspace{-0.3cm}
\end{table}

\textbf{Comparison with Shortcut-Reducing Data Augmentation.}
We further evaluate a heuristic data augmentation strategy that reduces regional shortcuts by emphasizing non-discriminative regions. As shown in Table~\ref{tab:shortcut_aug}, this strategy slightly improves the average accuracy over the baseline, indicating that reducing shortcut reliance can be beneficial. However, it remains clearly inferior to ARS-CDP, suggesting that explicit common-primitive modeling with dedicated constraints is more effective than heuristic input-level augmentation.

\begin{table}[H]
\centering
\vspace{-0.3cm}
\caption{Robustness comparison under primitive visibility perturbations on CIFAR-100 with ViT-B/16.}
\label{tab:robustness}
\resizebox{\columnwidth}{!}{
\begin{tabular}{lcccccccccc}
\toprule
Method & 0 & 1 & 2 & 3 & 4 & 5 & 6 & 7 & 8 & Avg. \\
\midrule
Baseline & 90.43 & 88.08 & 87.69 & 86.56 & 86.65 & 86.41 & 86.37 & 86.08 & 84.83 & 87.01 \\
+Add Gaussian Noise & 85.27 & 83.18 & 82.84 & 81.45 & 81.26 & 81.22 & 81.03 & 80.93 & 80.01 & 81.91 \\
+Corrupt Discriminative & 82.32 & 77.26 & 78.95 & 78.41 & 78.58 & 78.99 & 78.36 & 77.24 & 77.39 & 78.61 \\
\textbf{Ours} & \textbf{92.68} & \textbf{90.37} & \textbf{89.47} & \textbf{88.29} & \textbf{88.25} & \textbf{87.82} & \textbf{87.79} & \textbf{87.33} & \textbf{86.19} & \textbf{88.69} \\
+Add Gaussian Noise & 89.22 & 87.02 & 86.31 & 84.59 & 84.16 & 83.79 & 83.68 & 83.03 & 81.82 & 84.85 \\
+Corrupt Discriminative & 86.73 & 84.48 & 83.56 & 81.89 & 81.62 & 81.20 & 80.97 & 80.43 & 78.91 & 82.20 \\
\bottomrule
\end{tabular}
}
\vspace{-0.2cm}
\end{table}

\textbf{Robustness to Primitive Visibility Perturbations:}
We evaluate robustness under synthetic perturbations, including Gaussian noise and discriminative-region corruption. As shown in Table~\ref{tab:robustness}, both perturbations degrade performance, especially when discriminative regions are corrupted. However, ARS-CDP shows smaller performance drops than the baseline under the same perturbation settings, indicating that the proposed primitive modeling improves robustness to changes in primitive visibility.

\begin{table}[H]
\vspace{-0.3cm}
\centering
\caption{Computational overhead comparison with the baseline.}
\label{tab:overhead}
\resizebox{0.55\columnwidth}{!}{
\begin{tabular}{lcc}
\toprule
Method & Parameters & Time \\
\midrule
Baseline & 85.95M & 165.44s \\
Ours & 86.44M & 190.32s \\
\bottomrule
\end{tabular}
}
\vspace{-0.2cm}
\end{table}

\textbf{Computational Overhead:}
We compare the computational overhead of ARS-CDP with the baseline in terms of parameter count and per-epoch runtime. As shown in Table~\ref{tab:overhead}, ARS-CDP introduces only a small increase in parameters with a moderate runtime increase. This indicates that the additional overhead of the proposed primitive modeling is manageable.

\textbf{Limitations:}
ARS-CDP is mainly designed for standard image-based FSCIL settings, where the primitive structure can be derived from backbone-extracted visual features. Its generalization to more complex real-world multimedia scenarios, such as those involving multiple modalities or highly dynamic visual content, remains to be further investigated.

\section*{Acknowledgments}
This work is supported by the National Natural Science Foundation of China under grants 62206102; the National Key Research and Development Program of China under grant 2024YFC3307900; the National Natural Science Foundation of China under grants 62436003, 62376103 and 62302184; Major Science and Technology Project of Hubei Province under grant 2025BAB011 and 2024BAA008; Hubei Science and Technology Talent Service Project under grant 2024DJC078; Ant Group through CCF-Ant Research Fund. The computation is completed on the HPC Platform of Huazhong University of Science and Technology.

\bibliographystyle{ieeetr}
\bibliography{IEEEabrv, main}

\section{Biography Section}
\vspace{-1cm}
\begin{IEEEbiography}[{\includegraphics[width=1.2in,height=1.2in,clip,keepaspectratio]{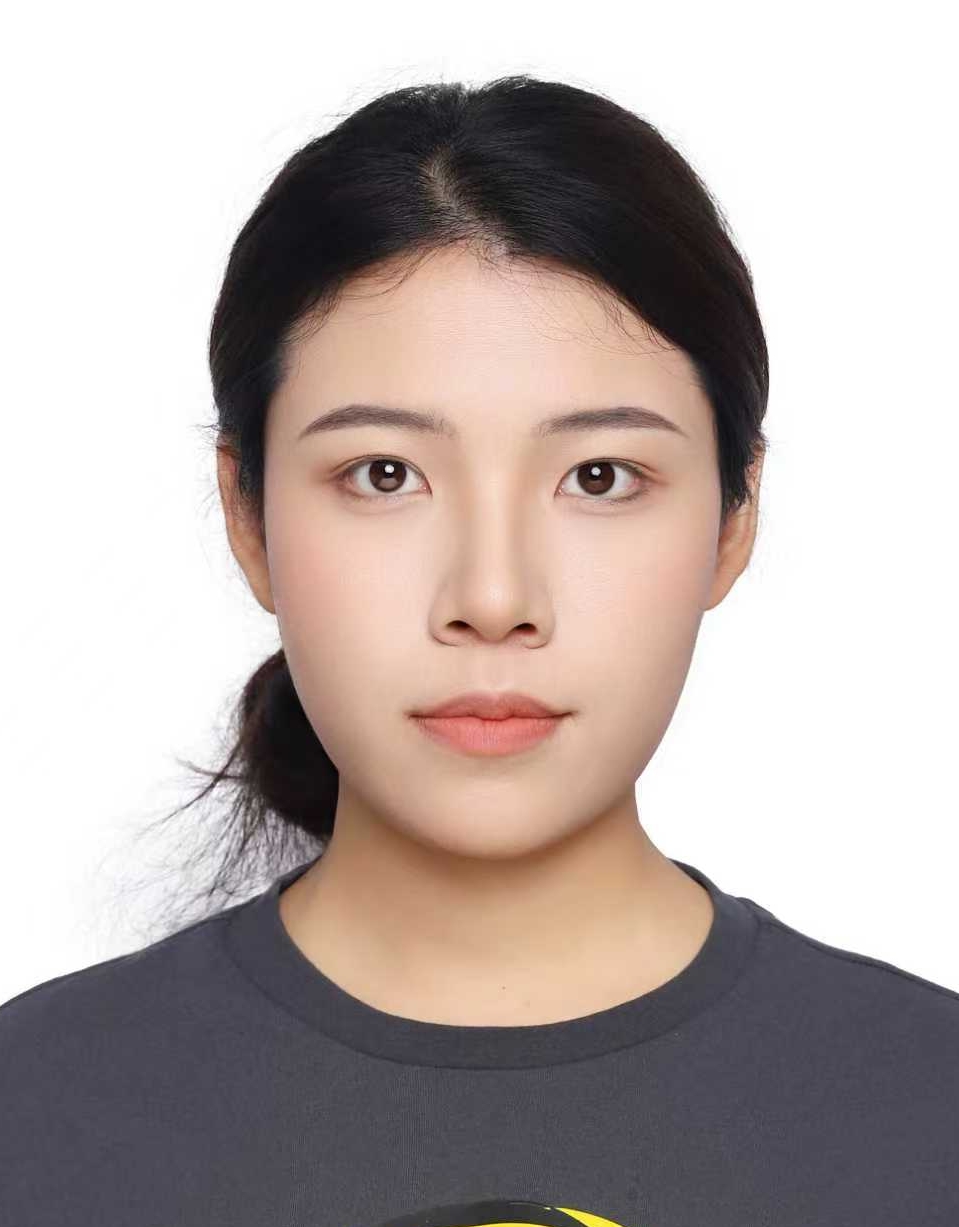}}]{Haichen Zhou} received the BS degree from the School of Management, Huazhong University of Science and Technology. She received the MS degree from the School of Computer Science and Technology, Huazhong University of Science and Technology. Her primary research interests include few-shot learning and incremental learning.
\end{IEEEbiography}
\vspace{-0.8cm}

\begin{IEEEbiography}[{\includegraphics[width=1.3in,height=1.3in,clip,keepaspectratio]{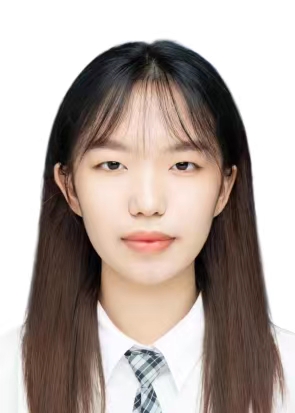}}]{Yazhe Lyu} received the BS degree from Harbin Engineering University, Harbin, China. She is currently working toward the master’s degree with the School of Computer Science and Technology, Huazhong University of Science and Technology. Her primary research interests include few-shot learning, zero-shot learning, and multimodal large language models.
\end{IEEEbiography}
\vspace{-0.8cm}
\begin{IEEEbiography}[{\includegraphics[width=1.2in,height=1.2in,clip,keepaspectratio]{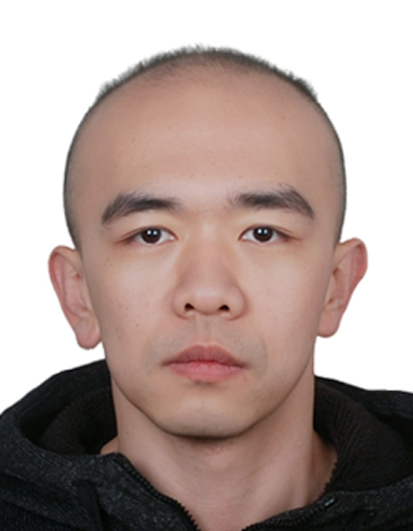}}]{Yixiong Zou} (Member, IEEE) 
received the B.S. degree from Nankai University, and received the Ph.D. degree from the School of Electrical Engineering and Computer Science, Peking University, Beijing, China. He was a visiting scholar at Carnegie Mellon University. He has published more than 40 journal and conference papers. He is currently a Lecturer with the School of Computer Science and Technology, Huazhong University of Science and Technology. His research interests include multimodal large language models, few-shot learning, open-world learning and computer vision.
\end{IEEEbiography}
\vspace{-0.8cm}

\begin{IEEEbiography}[{\includegraphics[width=1in, clip,keepaspectratio]{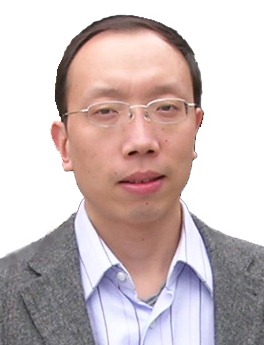}}]{Ruixuan Li} (Member, IEEE) received the B.S., M.S., and Ph.D. degrees in computer science from the Huazhong University of Science and Technology in 1997, 2000, and 2004, respectively. From 2009 to 2010, he was a Visiting Researcher with the Department of Electrical and Computer Engineering, University of Toronto. He is currently a Professor with the School of Computer Science and Technology, Huazhong University of Science and Technology. His research interests include cloud and edge computing, big data management, and distributed system security. He has published more than 500 journal and conference papers (NeurIPS, KDD, ICDM, IJCAI). He is also a member of ACM.
\end{IEEEbiography}
\vspace{-0.8cm}
\begin{IEEEbiography}[{\includegraphics[width=1.2in,height=1.2in,clip,keepaspectratio]{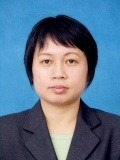}}]{Yuhua Li} (Member, IEEE) received the Ph.D. degree in computer application technology from Huazhong University of Science and Technology, Wuhan, China, in 2006. She is currently a Professor in the School of Computer Science and Technology, Huazhong University of Science and Technology. She was a visiting scholar at the University of California, Santa Barbara. She has published more than 60 journal and conference papers (NeurIPS, TKDE, SIGIR, WWW, ICDM, IJCAI). She is also a senior member of the China Computer Federation (CCF). Her research interests include data mining, social networks, machine learning, and big data.
\end{IEEEbiography}

\vfill

\end{document}